\documentclass[sigconf,nonacm]{acmart}

\usepackage[utf8]{inputenc} 
\usepackage[T1]{fontenc}    
\usepackage{url}            
\usepackage{booktabs}       
\usepackage{amsfonts}       
\usepackage{mathtools}      
\usepackage{nicefrac}       
\usepackage{microtype}      
\usepackage{xcolor}         
\usepackage{soul}           
\usepackage[normalem]{ulem}

\usepackage{amsmath}
\usepackage{mathtools}
\usepackage[noabbrev,capitalise]{cleveref}
\crefname{ineq}{Inequality}{Inequalities}
\creflabelformat{ineq}{#2{\upshape(#1)}#3}
\crefname{ALC@line}{line}{lines}
\Crefname{ALC@line}{Line}{Lines}
\crefname{ALC@unique}{line}{lines}
\Crefname{ALC@unique}{Line}{Lines}
\usepackage{algorithm}
\usepackage{algpseudocode}
\usepackage{bbm}
\usepackage{xspace}
\usepackage{subcaption}
\usepackage{natbib}

\newcommand{\R}{\mathbb{R}}
\newcommand{\cA}{\mathcal{A}}
\newcommand{\cX}{\mathcal{X}}
\newcommand{\cE}{\mathcal{E}}
\newcommand{\cD}{\mathcal{D}}
\newcommand{\cU}{\mathcal{U}}
\newcommand{\cL}{\mathcal{L}}
\newcommand{\cN}{\mathcal{N}}
\newcommand{\cB}{\mathcal{B}}
\newcommand{\cR}{\mathcal{R}}

\newcommand{\EE}{\mathbb{E}}
\newcommand{\Var}{\operatorname{\mathbb{V}}}

\newcommand{\mse}{\text{MSE}}
\newcommand{\Bias}{\text{Bias}}
\newcommand{\bias}{\text{bias}}
\newcommand{\var}{\text{var}}

\DeclarePairedDelimiter\abs{\lvert}{\rvert}%

\newcommand{\vmips}{\hat v_{\text{MIPS}}}
\newcommand{\vips}{\hat v_{\text{IPS}}}

\theoremstyle{plain}
\newtheorem{assumption}{Assumption}

\newtheorem{proposition}{Proposition}

\newcommand{\method}{CAEL-MIPS\xspace}

\title{Context-Action Embedding Learning for Off-Policy Evaluation in Contextual Bandits}

\author{Kushagra Chandak}
\authornote{Work done during an internship at RBC Borealis.}
\affiliation{%
  \institution{RBC Borealis}
  \country{}
  \city{}
}
\email{kchandak@ualberta.ca}

\author{Vincent Liu}
\affiliation{%
  \institution{RBC Borealis}
  \country{}
  \city{}
}
\email{vincent.liu@rbc.com}

\author{Haanvid Lee}
\affiliation{%
  \institution{RBC Borealis}
  \country{}
  \city{}
}
\email{haanvid.lee@rbc.com}

\begin{document}

\begin{abstract}
    We consider off-policy evaluation (OPE) in contextual bandits with finite action space. Inverse Propensity Score (IPS) weighting is a widely used method for OPE due to its unbiasedness, but it suffers from significant variance when the action space is large or when some parts of the context-action space are underexplored. Recently introduced Marginalized IPS (MIPS) estimators mitigate this issue by leveraging action embeddings. However, these embeddings neither minimize the mean squared error (MSE) of the estimators nor incorporate contextual information. To address these limitations, we introduce Context-Action Embedding Learning for MIPS (CAEL-MIPS), which learns context-action embeddings from offline data to minimize the MSE of the MIPS estimator. Building on the theoretical analysis of bias and variance of MIPS, we present an MSE-minimizing objective for CAEL-MIPS and show its connection to information-theoretic quantities. In the empirical studies on a synthetic dataset and a real-world dataset, we demonstrate that our estimator outperforms baselines in terms of MSE.
\end{abstract}



\maketitle

\section{Introduction}\label{sec:intro}

The contextual bandit framework \citep{lattimore2020bandit} has been applied to various web services such as recommendation systems \citep{li2010contextual, xu2020contextual} and digital marketing \citep{geng2021comparison, thomas2017predictive}. In this framework, the system selects an action based on contextual information and receives a reward for that action. For example, in news article recommendation, the system recommends a news article (action) based on the user's interests (context), and receives a reward if the user clicks on the article. Since such systems are updated frequently, a key challenge is evaluating the performance of each update. The most common approach is to perform an A/B test, that is, to deploy both the updated system and the baseline in an online experiment. However, A/B testing can be costly in practice, as the updated system can lead to poor user experience or revenue loss. Moreover, A/B tests are often slow, with final evaluations taking several weeks or even months, thereby delaying system updates and progress.

In the off-policy evaluation (OPE) problem, the goal is to evaluate the performance of an algorithm or a \emph{policy} using historical data. In applications such as recommendation systems and digital marketing, we often have logs of historical data, and OPE can be used to evaluate the performance of a \emph{target} policy using the logged data directly \citep{dudik2011doubly,lee2022local}. This offline evaluation approach alleviates the need for A/B testing, enabling faster progress and improving user experience.

Most OPE estimators are based on Inverse Propensity Score (IPS) weighting \citep{horvitz1952generalization,dr}, often called importance sampling \citep{sutton1998reinforcement}. IPS corrects the sampling distribution to the target distribution by applying IPS weights. It is unbiased under certain conditions, and has been proven to be optimal for contextual bandits \citep{wang2017optimal}. Despite such desired theoretical properties, estimators based on IPS suffer from high variance in practice, especially when there is a large mismatch between the target policy and the data collection policy, or when the number of actions is large. 


To address the high variance problem of IPS, \citet{saito2022off} recently proposed the Marginalized IPS (MIPS) estimator. 
MIPS applies IPS weighting in an embedding space. Typically, the embedding space has a smaller dimension compared to the action space, which helps reduce the variance of IPS. \citet{saito2022off} assume that the embeddings are available to the estimator, and theoretically characterize how the embeddings can be used for bias-variance trade-off.
For example, reducing the embedding dimension can reduce variance at the cost of increasing bias. 

\citet{cief2024learning} extend the work of \citet{saito2022off} by learning action embeddings using a linear reward model with a reward prediction loss and using the last layer of the model as embeddings. But their method learns action embeddings that are not conditioned on contexts. As a result, the action embeddings for different contexts can be similar even when the corresponding rewards are very different. For example, in clinical trials, a particular dose of adrenaline (action) can be life-saving or lethal depending on the patient's condition (context) \citep{nordseth2012dynamic}. Learning only action embeddings will learn the same embedding for an adrenaline dosage regardless of the patient's condition. Moreover, it is unclear whether only minimizing the reward prediction loss is sufficient to learn good embeddings. In this work, we address these issues by proposing a novel objective to learn embeddings that are conditioned on both actions and contexts, and which directly minimizes the mean squared error (MSE) of MIPS. \citet{kiyohara2024off} also proposes a similar embedding learning objective, but for the slate bandit model, where each slate contains multiple actions. While their objective intuitively minimizes bias and variance of the estimator, it is unclear if the learned embeddings actually do so. In contrast, our objective directly estimates bias and variance, yielding MSE-minimizing embeddings.

Several OPE estimators use embeddings or similar notions to reduce variance at the cost of a small bias. \citet{sachdeva2024off} perform policy convolution via a similarity function between action embeddings with the implicit assumption that similar embeddings lead to similar rewards. 
\citet{olivares2025clustering} directly partition the context space into clusters and applies IPS over clusters instead of contexts, based on the assumption that the context does not affect the reward given the action and the cluster. Our work is different since our goal is to learn good embeddings for OPE, and our proposed embedding learning method can be potentially extended to these estimators. 

In this paper, we propose learning context-action embeddings for the MIPS estimator by minimizing its mean squared error (MSE). In our theoretical studies, we first derive upper bounds on the bias and variance of MIPS, and propose an embedding learning objective using the upper bounds. Our proposed objective learns embeddings by balancing the bias and variance of MIPS so that the resulting estimator has a smaller MSE compared to the previous work of \citet{cief2024learning}. We also show information-theoretic connections to our objective, and how it grows as the number of actions increases. In our empirical studies, we conduct experiments on a synthetic dataset and a real-world bandit dataset collected from a fashion e-commerce platform \citep{saito2020large}. In both sets of experiments, we demonstrate that our method further reduces the MSE of the estimator proposed in \citet{cief2024learning}, which only considers reward prediction for embedding learning. 

\section{Preliminaries}\label{sec:setting}
In this section, we introduce the problem setting for OPE in contextual bandits, and MIPS estimator upon which we build our proposed method.

\paragraph{Notation.} Random variables are denoted with uppercase letters in standard font, such as $X, Y, Z$, etc. Expected value $\EE$ is over the data-generating distribution unless specified otherwise. The notation $[k]$ denotes the set $\{1,\dots,k\}$.

\subsection{Off-policy Evaluation in Contextual Bandits}\label{sec:ope-bandits}
Let $\cX \subseteq \R^d$ be the context space, $\cA$ be a finite set of actions, and $\cR$ be the space of rewards. Data are collected offline by following a \emph{behavior policy} $\mu \colon \cX \to \Delta(\cA)$ in the following way. First, a context $X \sim p_X (\cdot)$ is sampled where $p_X$ is the unknown context distribution. Then an action $A \sim \mu(\cdot | X)$ is sampled. Finally, a reward $R \sim p_R{(\cdot|X,A)}$ is received, where $p_{R}$ is the unknown reward distribution given context $X$ and action $A$. The reward $R$ is the \emph{bandit feedback} as it only reveals the value of the action taken in a context. We collect such iid samples in a dataset $\cD = ((X_i,A_i,R_i))_{i=1}^n$ of size $n$.

For a policy $\pi \colon \cX \to \Delta(\cA)$, let
\begin{align*}
    v(\pi) \coloneq \int_{\cX} \sum_{a \in \cA} p_X(x) \pi(a|x) q(x,a) dx \,.
\end{align*}
Here $q \colon \cX \times \cA \to \R$ is the mean reward function $q(x,a) \coloneq \int r p_R(r|x,a) dr$.

The goal in OPE is to estimate $v(\pi)$ for a \emph{target policy} $\pi$ using the offline dataset $\cD$. The performance of an estimator $\hat v(\pi)$ is measured using the mean squared error or the MSE:
\begin{align*}
    \mse[\hat v(\pi)] &\coloneq \EE[(\hat v(\pi) - v(\pi))^2] \\
    &=\Bias[\hat v(\pi)]^2 + \Var[\hat v(\pi)]\,,
\end{align*}
where $\Bias[\hat v(\pi)]$ and $\Var[\hat v(\pi)]$ denote the bias and variance of the estimator $\hat v(\pi)$ respectively.

\subsection{Marginalized IPS Estimator}\label{sec:related-work}
Perhaps the most widely used estimator for OPE is the Inverse Propensity Score (IPS) weighting, sometimes referred as importance sampling. Given a behavior policy $\mu$, a target policy $\pi$, and a dataset $\cD = ((X_i, A_i, R_i))_{i=1}^n$, the IPS estimator estimates $v(\pi)$ as
\begin{align*}
    \vips(\pi) \coloneq \frac1n \sum_{i=1}^n \underbrace{\frac{\pi(A_i|X_i)}{\mu(A_i|X_i)}}_{=:w(X_i,A_i)} R_i\,,
\end{align*}



where $w(x,a)$ is the IPS weight for $(x,a) \in \cX\times\cA$. Under the following coverage assumption on the action space, the IPS estimator is unbiased, i.e., $\EE[\vips(\pi)] = v(\pi)$. 

\begin{assumption}[Common support]\label{as:common-supp}
     $\pi(a|x) > 0 \implies \mu(a|x) > 0$ for all $x \in \cX$ and $a \in \cA$.
\end{assumption}
In practice \cref{as:common-supp} is often violated, making IPS biased. Under such conditions, the bias of IPS is given by \citep{saito2022off, sachdeva2024off}:
\begin{align*}
    |\Bias[\vips(\pi)]| = \EE\left[ \sum_{a \in \cU(X,\mu)} \pi(a|X) q(X,a) \right] \,,
\end{align*}
where $\cU(x, \mu) = \{ a \in \cA : \mu(a|x) = 0 \}$ is the set of unsupported actions in context $x$ under policy $\mu$. A bigger issue is the variance of IPS \citep{dr, saito2022off}, given by
\begin{align*}
     \Var[\vips(\pi)] =& \frac1n\EE\!\left[ w(X,A)^2 \Var[R|X,A] \right] + \Var\left[ \EE_\mu[w(X,A) q(X,A)] \right]  \\
     & + \EE\left[ \Var_\mu[w(X,A) q(X,A)] \right]\,,
\end{align*}
where $\EE_\mu$ and $\Var_\mu$ are expectation and variance respectively under the behavior policy $\mu$. The above equation shows that the variance of IPS is high when the IPS weights $w$ are large or when $\Var[R|X,A]$ is large, i.e., the rewards are noisy.

To alleviate the issues of IPS, \citet{saito2022off} introduced the Marginalized IPS (MIPS) estimator. The MIPS estimator assumes access to embeddings from some set $\cE \subseteq \R^{d_\cE}$. The embeddings are sampled from an unknown embedding distribution $p_E{(e|x,a)}$. The rewards are assumed to be sampled from $p_R{(r|x,a,e)}$. Now the value of a policy is defined as
\begin{align*}
    v(\pi) \coloneq \int_\cX \sum_{a\in \cA} \int_\cE p_X(x) \pi(a|x)p_E(e|x,a) q(x,a,e) dx de \,
\end{align*}
where $q(x,a,e) \coloneq \int r p_R(r|x,a,e)dr$ is the reward function or expected reward given context $x$, action $a$, embedding $e$. The \emph{marginal} embedding distribution induced by a policy $\pi$ is defined as $p_E(e|x,\pi) \coloneq \sum_{a \in \cA} p_E(e|x,a) \pi(a|x)$. The goal is still to estimate $v(\pi)$ using samples. Given dataset $\cD \coloneq ((X_i, A_i, E_i, R_i))_{i=1}^n$, the MIPS estimator \citep{saito2022off} is defined as
\begin{align}
    \vmips(\pi) &\coloneq \frac1n \sum_{i=1}^n \underbrace{ \frac{p_E(E_i|X_i, \pi)}{p_E(E_i|X_i, \mu)} }_{_{=:\rho(X_i,E_i)}} R_i  \label{eq:mips}\\
    &= \frac1n \sum_{i=1}^n \sum_{a\in\mathcal{A}}  \underbrace{\frac{\mu(a|X_i)  p_E(E_i|X_i,a)}{p_E(E_i|X_i,\mu)}}_{=:\mu(a|X_i,E_i)}w(X_i,a)R_i \label{eq:mips_in_practice}
\end{align}


where $\rho(X,E)$ is the marginal IPS weight for a pair of context and embedding $(X,E)$ in the offline data, and with some overload of notation, $\mu(a|X,E)$ is the posterior distribution of action $a\in\mathcal{A}$ given $(X,E)$ in the offline data \citep{saito2022off}. In practice, the estimator $\vmips(\pi)$ is computed following \cref{eq:mips_in_practice}, where the posterior $\mu(a|X,E)$ is estimated via a logistic regression model \citep{saito2020large}. MIPS requires a common support assumption on embeddings similar to \cref{as:common-supp} for having finite marginal IPS weights. 

\begin{assumption}[Common Embedding Support, Assumption 3.1 in~\citep{saito2022off}]\label{as:common-embed-supp}
    Given policies $\pi$ and $\mu$, $p(e|x,\pi) > 0 \implies p(e|x,\mu) > 0$ for all $e \in \cE$ and $x \in \cX$, where $p_E(e|x,\pi)$ is the marginal embedding distribution given policy $\pi$ and $x \in \cX$ defined as $p(e|x,\pi) \coloneq \sum_{a \in \cA} p_E(e|x,a) \pi(a|x)$.
\end{assumption}
Note that the embedding space is generally smaller than the action space to reduce the variance of the MIPS estimator. Therefore, this assumption is weaker than the common support assumption (\cref{as:common-supp}). Under the common embedding support assumption (\cref{as:common-embed-supp}), \citet{saito2020large} showed that the bias of MIPS can be decomposed as
\begin{align}
    & \Bias[\vmips(\pi)] = \EE\Biggl[ \sum_{a < b} \mu(a|X,E)\mu(b|X,E) \cdot \nonumber \\
    & \qquad (q(X,a,E) - q(X,b,E)) \cdot (w(X,b) - w(X,a)) \Biggr] \label{eq:mips-bias} \,.
\end{align}

The MIPS estimator is unbiased under the following assumption.

\begin{assumption}[No Direct Effect, Assumption 3.2 in~\citep{saito2022off}] \label{as:no-direct-eff}
    Action $A$ is said to have no direct effect on reward $R$ if they are conditionally independent given the context $X$ and the embedding $E$: $A \perp R \, | \, X,E$.
\end{assumption}
If, in addition to \cref{as:common-embed-supp}, \cref{as:no-direct-eff} also holds, then the bias in \cref{eq:mips-bias} becomes zero.” due to $q(X,a,E)=q(X,b,E)$. \citet{saito2022off} also showed that under Assumptions \ref{as:common-supp}, \ref{as:common-embed-supp}, and \ref{as:no-direct-eff}, the variance reduction of MIPS compared to IPS is given by
\begin{align}
    & \Var[\vips(\pi)] - \Var[\vmips(\pi)] \nonumber\\
    &= \frac{1}{n}\EE[ \EE_{p_R{(\cdot|X,E)}}[R^2]\Var_{\mu(\cdot |X,E)}[w(X,A)]] \label{eq:mips-var0} \,,
\end{align}
where $p_R(\cdot|X,E)$ is the reward distribution that depends only on $X$ and $E$ and not on $A$.

From \cref{eq:mips-bias}, note that the bias is small when the posterior probabilities $\mu(a|x,e)$ are close one or zero. However, in such cases, \cref{eq:mips-var0} shows that the variance reduction is small, or the variance of MIPS estimator increases closer to the variance of IPS estimator, since the term $\Var_{\mu(\cdot \,|\, X,E)}[w(X,A)]$ is small due to reduced stochasticity of the posterior probability.

However, the MIPS estimator does not have control over its bias and variance as it assumes that the embeddings are given. In real-world applications, embeddings are not given in many cases. Even if the embeddings are provided, they might not be optimal embeddings that minimize the MSE of the MIPS estimator, as they were not designed to do so.

To learn embeddings based on offline data, \citet{cief2024learning} proposed an embedding learning method for MIPS that learns action embeddings by minimizing a reward prediction loss. However, embeddings that minimize the reward prediction loss do not necessarily minimize the MSE of the MIPS estimator. Specifically, their embedding learning objective does not account for how concentrated or spread out the posterior $\mu(a \,|\, x, e)$ is, where the posterior controls the bias and variance of MIPS estimator as shown in \cref{eq:mips-bias,eq:mips-var0}. Moreover, the work of \citet{cief2024learning} only learns action embeddings, while the embeddings assigned to actions should be different depending on the context, as an action may have a different reward in different contexts. 

In the following section, we introduce our proposed method \textbf{C}ontext-\textbf{A}ction \textbf{E}mbedding \textbf{L}earning for \textbf{MIPS} (\method) that learns MSE-minimizing embeddings that depend on both contexts and actions.

\section{Context-Action Embedding Learning}\label{sec:method}
\method learns context-action embeddings by minimizing bias and variance of MIPS. To this end, we derive upper bounds for bias and variance and propose an embedding learning objective that combines the reward prediction error from \citet{cief2024learning} with these upper bounds. The proofs for all the results are deferred to Appendix~\ref{appendix:proofs}.

\subsection{Embedding Learning through MIPS MSE Minimization}

As the bias and variance of MIPS cannot be directly evaluated from offline data, we derive their upper bounds.

\subsubsection{Upper bound of MIPS Bias}
The bias of MIPS (\cref{eq:mips-bias}) cannot be evaluated using offline data as we cannot estimate the mean reward $q(x,a,e) = \int r p_R(r|x,a,e)$ since we do not have reward sampled from $p_R(r|x,a,e)$. Therefore, we derive an upper bound of bias that can be evaluated with offline data under the following mild assumption.

\begin{assumption}\label{as:bounded-mean-var}
    The mean and the variance of the rewards are bounded:
    \begin{align*}
        & q_{\min} \le q(x,a,e) \le q_{\max}\,, \\
        & \sigma^2_{\min} \le \Var[R \,|\,X=x, E=e] \le \sigma^2_{\max} \,,        
    \end{align*}
    for all $(x,a,e) \in \cX \times \cA \times \cE$.
\end{assumption}
Given that the mean and the variance are bounded in most applications, e.g. click through rate of a webpage is between 0 and 1, \cref{as:bounded-mean-var} is generally satisfied in real-world scenarios. The following theorem establishes an upper bound on the bias of the MIPS estimator.
\begin{proposition}\label{prop:bias-bound}
Under Assumptions \ref{as:common-embed-supp} and \ref{as:bounded-mean-var}, the upper bound of MIPS estimator's bias is given by 
    \begin{align*}
        & \Bias[\vmips(\pi)]^2 \\
        & \le  (q_{\max}-q_{\min})^2  \EE\Biggl[ \sum_{a < b} \mu(a|X,E)\mu(b|X,E)  \abs{w(X,b) - w(X,a)} \Biggr]^2. 
    \end{align*}
\end{proposition}
To evaluate the upper bound in \cref{prop:bias-bound} with offline data, we use a function approximator $f_\theta$ to obtain context-action embeddings $E \coloneq f_\theta(X,A)$. Using the embeddings, we estimate the posterior $\mu(\cdot \,|\,X,f_\theta(X,A))$ as in the MIPS estimator via logistic regression. Finally, we use the estimated posterior and the IPS weights $w(\cdot \,, \cdot)$ to evaluate the upper bound in \cref{prop:bias-bound}. Since the exact bounds on the mean rewards, $q_{\max}$ and $q_{\min}$, are typically unknown, we introduce a hyperparameter $\alpha$ in our final objective (see \cref{eq:final-obj}). To learn bias-minimizing embeddings from offline data, we formulate the bias upper bound as an objective $\cL_{\text{bias}}(\theta)$ for \method:
\begin{align}
\label{eq:bias-loss}
    \cL_{\text{bias}}(\theta) & \coloneq \EE\Biggl[  \sum_{a<b} {\mu(a|X, f_\theta(X,A))\mu(b|X, f_\theta(X,A))} \cdot \notag \\
    & \qquad \qquad \abs{w(X,b) - w(X,a)} \Biggr]^2 \,.
\end{align}
From \cref{eq:bias-loss}, we can see that the bias loss term approaches zero as the stochasticity of the posterior $\mu(a \,|\, X,f_{\theta}(X,A))$ is reduced ($\mu(a \,|\, X,f_{\theta}(X,A)) \to 0$ or 1). The embedding learning function $f_\theta (X, A)$ that minimizes this loss would induce the posterior probability closer to 1 or 0 when the difference of IPS weights $|w(X,b) - w(X,a)|$ is large due to large discrepancy between $\pi$ and $\mu$ on context $X$ sampled in the offline data.

\subsubsection{Upper bound of MIPS Variance} The variance reduction of MIPS compared to IPS is given by \cref{eq:mips-var0}. In order to evaluate the variance of MIPS using offline data without calculating the variance of the IPS weights, we establish an upper bound on the variance of MIPS in the next proposition.
\begin{proposition}\label{prop:var-bound}
    When Assumptions \ref{as:common-supp}, \ref{as:common-embed-supp}, \ref{as:no-direct-eff}, and \ref{as:bounded-mean-var} hold, then
    \begin{align*}
        & \Var[\vmips(\pi)] \le \Var[\vips(\pi)] \\ 
         & + \frac{1}{n}\EE\left[ \left(\sigma^2_{\min} + \EE[R|X,E]^2 \right) \sum_{a \in \cA} \mu(a|X,E)^2 \sum_{a \in \cA} w(X,a)^2 \right] \,.
    \end{align*}
\end{proposition}
Similar to \cref{prop:bias-bound}, we evaluate the upper bound in \cref{prop:var-bound} by parametrizing $E = f_\theta(X,A)$ and estimating the posterior $\mu(\cdot \,|\, X, f_\theta(X,A))$ using offline data. Using $E$ and $X$, we use a reward prediction model to get $E[R|X,E]^2$. Finally, with the estimated posterior, the IPS weights, and $E[R|X,E]^2$, we can evaluate the upper bound in \cref{prop:var-bound}. Furthermore, since the first term $\Var[\vips(\pi)]$ in the upper bound in \cref{prop:var-bound} does not depend on the embeddings, we can ignore it while minimizing the variance upper bound to learn embeddings. We introduce another hyperparameter $\beta$ in the final objective (see \cref{eq:final-obj}) since $\sigma_{\min}^2$ is not typically known in advance. We define the objective based on variance as
\begin{align}\label{eq:var-loss}
    \cL_{\text{var}}(\theta) \coloneq & \frac{1}{n} \EE\Bigg[ \EE[R\,|\,X,E]^2 \sum_{a \in \cA}  \mu (a|X, f_\theta(X, A))^2 \nonumber \\
    & \qquad \qquad\sum_{a \in \cA} w(a, X)^2 \Bigg]
\end{align}

Note that $\cL_{\text{var}}(\theta) \to 0$ when $\sum_a \mu(a \,|\, X, f_\theta(X,A))^2$ is minimized. This sum of squared probabilities is minimized when the posterior $\mu(\cdot\,|\, X, f_\theta(X,A))$ is a uniform distribution over the actions (see \cref{prop:entropy-var}). Learning the embedding function $f_\theta$ to minimize the upper bound of variance $\cL_{\text{var}}(\theta)$ (\cref{eq:var-loss}) induces the posterior to be more uniform on the context $X$ where $\sum_{a \in \cA} w(a, X)^2$ is high. In other words, when there is high variance due to a large discrepancy between the behavior and target policies \citep{saito2022off, dr}, the embedding function is learned to smooth out the induced posterior distribution. Also note that $\cL_\var(\theta) \to 0$ as $n \to \infty$. This is expected as the variance of an estimator goes to zero as the sample size goes to infinity. So minimizing \cref{eq:var-loss} is more useful for small sample sizes.

\subsubsection{Embedding Learning Objective}
\cref{prop:bias-bound,prop:var-bound} show the upper bounds of bias and variance of MIPS. On one hand, by learning the embeddings to minimize the upper bound of bias, $\mu(\cdot|X,f_\theta(X,A))$ induced from the learned embeddings concentrates around some action as discussed earlier. On the other hand, minimizing upper bound of variance pushes the posterior distribution towards a uniform distribution. By balancing this trade-off, we propose an embedding learning objective that combines the upper bounds of bias and variance, as well as the reward prediction loss introduced by \citet{cief2024learning}. The reward prediction loss is used since we would like to learn embeddings that capture most of the reward information from contexts and actions to further reduce bias. It is also empirically shown to have good performance in previous works \citep{cief2024learning,kiyohara2024off}. Our final objective function is
\begin{align}\label{eq:final-obj}
    \cL(\theta) \coloneq \cL_R(\theta) + \alpha\cL_{\text{bias}}(\theta) + \beta\cL_{\text{var}}(\theta) \,,
\end{align}
where $\cL_R(\theta)$ is the MSE of reward prediction and $\alpha$ and $\beta$ are positive coefficients. The first term $\cL_R(\theta)$ in \cref{eq:final-obj} aims to reduce bias since predicting the rewards well based on embeddings controls bias as seen in \cref{eq:mips-bias} and \cref{prop:bias-bound}. The second and the third term aim to reduce bias and variance respectively, as governed by the posterior distribution $\mu$ and discussed in \cref{prop:bias-bound,prop:var-bound}. The hyperparameters $\alpha$ and $\beta$ control the contribution of $\cL_\bias(\theta)$ and $\cL_\var(\theta)$ to the total loss. Larger values of $\alpha$ and $\beta$ induce larger bias and variance reduction respectively. We estimate \cref{eq:final-obj} using offline data $\cD$ by replacing expectations with empirical averages to get $\hat \cL(\theta; \cD) \coloneq \hat\cL_R(\theta; \cD) + \alpha\hat\cL_{\text{bias}}(\theta; \cD) + \beta\hat\cL_{\text{var}}(\theta; \cD)$. Recall that $f_\theta(x,a)$ outputs the embedding of $(x,a) \in \cX\times\cA$, which is used for reward prediction $\hat R$. The reward prediction loss, the bias loss, and the variance loss that consist the embedding learning loss $\hat \cL(\theta; \cD)$ are:
\begin{align}
    & \hat \cL_R(\theta; \cD) \coloneq \frac1n \sum_{i=1}^n (\hat R(X_i, f_\theta(X_i,A_i)) - R_i)^2  \,, \label{eq:emp-r-loss}
\end{align}
\begin{align}
    &\hat \cL_{\text{bias}}(\theta; \cD) \nonumber \\
    & \coloneq \frac{1}{n^2} \Biggl(\sum_{i=1}^n \sum_{a < b} {\hat\mu(a|X_i, f_\theta(X_i,A_i))\hat\mu(b|X_i, f_\theta(X_i,A_i))} \cdot \nonumber \\
    & \qquad \qquad \qquad \abs{w(X_i,b) - w(X_i,a)} \Biggr)^2  \,, \label{eq:emp-b-loss}
\end{align}
and
\begin{align}
    & \hat \cL_{\text{var}}(\theta; \cD) \coloneq \frac{1}{n^2} \sum_{i=1}^n \hat R(X_i, f_\theta(X_i,A_i))^2 \sum_{a \in \cA} \hat\mu(a|X_i, f_\theta(X_i, A_i))^2 \cdot \nonumber \\
    & \qquad \qquad \qquad \sum_{a \in \cA}  w(a, X_i)^2 \label{eq:emp-v-loss} \,.
\end{align}


\subsubsection{Discussion on the Bias and Variance Upper Bounds}
We give some intuitions on the bias and variance upper bounds via the entropy of the posterior, and show their growth with respect to the size of the action space. Let 
\begin{align*}
    H_2(X,A; \theta) = - \log\sum_{a \in \cA} \mu(a|X,f_\theta(X,A))^2
\end{align*}
be the R\'{e}nyi entropy of order 2 of the posterior $\mu(a|X,f_\theta(X,A))$. The next proposition relates the bias upper bound with the entropy of the posterior and the size of the action set.
\begin{proposition}\label{prop:entropy-bias}
    Let $C > 0$ and $|w(x,a) - w(x,b)| < C$ for all $x \in \cX$, and for all $a,b \in \cA$. Then
    \begin{align*}
        \cL_\bias(\theta) \le \frac{C^2}{4} \left( 1- \EE\left[ e^{-H_2(X, A; \theta)} \right] \right)^2 \le \frac{C^2}{4}\left( 1 - \frac{1}{|\cA|} \right) \,,
    \end{align*}
    where $H_2(X, a;\theta)$ is the R\'{e}nyi entropy of order 2 of the posterior $\mu(\cdot |X,f_\theta(X,A))$. Further, $\cL_\bias(\theta) \le (C^2/4)\EE[H_2(X,A;\theta)]$.
\end{proposition}

\cref{prop:entropy-bias} states that minimizing the entropy $H_2$ corresponds to minimizing the bias upper bound. The entropy is minimized when $\mu(\cdot \,|\, X,f_\theta(X,A))$ is concentrated around some action. This means that when the context $X$ and the learned embedding $f_\theta(X,A)$ can fully recover an action, then the bias upper bound is zero. This verifies our earlier intuition about $\cL_\bias(\theta)$. From \cref{prop:entropy-bias}, we also observe that as the number of actions increases, the bias upper bound increases and is further upper bounded by $(1-1/|\cA|)$. Later in \cref{sec:synthetic-exp}, we validate this observation empirically. Next, we show that minimizing the variance upper bound is equivalent to maximizing $H_2$ and show the convergence rate of the variance upper bound with respect to the dataset size and the number of actions.

\begin{proposition}\label{prop:entropy-var}
    Let $H_2(X, A; \theta)$ be the R\'{e}nyi entropy of order 2. Then
    \begin{align*}
        \cL_\var(\theta) = \frac1n \EE\left[ \EE[R|X,E]^2 \cdot e^{-H_2(X,A;\theta)} \sum_{a \in \cA} w(X, a)^2 \right] \,. 
    \end{align*}
    Further, we have that $\cL_\var(\theta) \le \frac1n \sum_a\EE[\EE[R|X,E]^2 w(X,a)^2] = O(|\cA|/n)$ for bounded IPS weights.
\end{proposition}

From \cref{prop:entropy-var}, we observe that minimizing the variance loss is equivalent to maximizing $H_2$, weighted by the sum of squared IPS weights. The entropy $H_2$ is maximized when the posterior $\mu(\cdot \,|\, X,f_\theta(X,A))$ is uniform implying that variance minimization encourages embeddings that yield a uniform posterior over actions. The weighting emphasizes regions where the squared IPS weights are large, that is, where the overlap between the behavior and target policies is small. Additionally, we observe that the variance loss increases linearly with the number of actions, which we verify empirically in \cref{sec:synthetic-exp}.

\subsection{Algorithm}\label{sec:alg}
The pseudocode for \method is presented in \cref{alg:main-algo}. \method uses a three-layer feedforward neural network $f_\theta$ to parameterize the embeddings and the rewards. The hidden layers use batchnorm, ReLU activations, and a dropout rate of 0.2. The network takes $(x,a) \in \cX \times \cA$ as input and outputs the embedding $\hat e$ for the context $x$ and action $a$, that is, $\hat e \coloneq f_\theta(x,a)$. Following \citet{cief2024learning}, the network also outputs a reward prediction using the embeddings given by $\hat r \coloneq \hat e^\top x$.

\begin{algorithm}[t]
\caption{Computing $\vmips(\pi)$ using \method}
\label{alg:main-algo}
\begin{algorithmic}[1]
  \State \textbf{Input:} Offline dataset $\cD = ((X_i,A_i,R_i))_{i=1}^n$, behavior policy $\mu$, target policy $\pi$, embedding network $f_\theta(X,A)$ with parameters $\theta$, posterior with $\hat\mu_\phi(\cdot \,|\, X,E)$ parameters $\phi$, batch size $B$, learning rate $\eta$.

  \For {each iteration}
      \State Sample mini-batch $\cB = ((X_i,A_i,R_i))_{i=1}^B$ from $\cD$ \label{line:mini-batch}
      \State Compute $\hat E_i \gets f_\theta(X_i, A_i)$  \label{line:e}
      \State Fit $\hat \mu_\phi(\cdot |X_i, \hat E_i)$ by supervised learning on $\cB$ \label{line:log-regr}
      \State Compute $\hat R_i \gets \hat E_i^\top X_i$  \label{line:r-e}
      \State Compute $\hat \cL(\theta; \cB)$ using $\hat E_i$, $\hat R_i$, $\hat{\mu}_\phi (\cdot | X_i, A_i)$, $\pi$, and $\mu$ (\cref{eq:emp-r-loss,eq:emp-b-loss,eq:emp-v-loss}) \label{line:total-loss}
      \State Update $\theta$ \label{line:grad-step}
      \begin{equation}
        \theta \leftarrow \theta -\eta\nabla_\theta \hat \cL(\theta; \cB) \notag  
      \end{equation}
  \EndFor
  \State Fit $\hat \mu_\phi(\cdot | X_i, \hat E_i)$ by supervised learning on $((\hat E_i, A_i))_{i=1}^n$ \label{line:log-regr2}
  \State Compute the MIPS estimator: \label{line:mips}
  \begin{equation}
      \vmips \gets \frac{1}{n}\sum_{i=1}^n \sum_{a \in \cA} \hat \mu_\phi (a|X_i,f_\theta(X_i, A_i)) w(X_i, a) R_i \notag 
  \end{equation}
  \Return $\vmips$
\end{algorithmic}
\end{algorithm}

To compute $\vmips(\pi)$, \cref{alg:main-algo} first learns the context-action embeddings $(\hat E_i)_{i=1}^n$ by performing gradient updates on the loss defined in \cref{eq:final-obj} (lines~\ref{line:mini-batch}-\ref{line:grad-step}). After learning the embeddings, \cref{alg:main-algo} estimates $\hat \mu_\phi$ using supervised learning on $((\hat E_i, A_i))_{i=1}^n$ (line~\ref{line:log-regr2}). We use logistic regression from \emph{scikit-learn} \citep{pedregosa2011scikit} to estimate $\hat \mu_\phi$ following \citet{saito2020large} and \citet{cief2024learning}. Finally, on line~\ref{line:mips}, \cref{alg:main-algo} computes the MIPS estimator using $\hat \mu_\phi$ as in \cref{eq:mips_in_practice}.

\section{Experiments}\label{sec:experiments}

\begin{figure}[!ht]
    \centering
    \includegraphics[width=\linewidth]{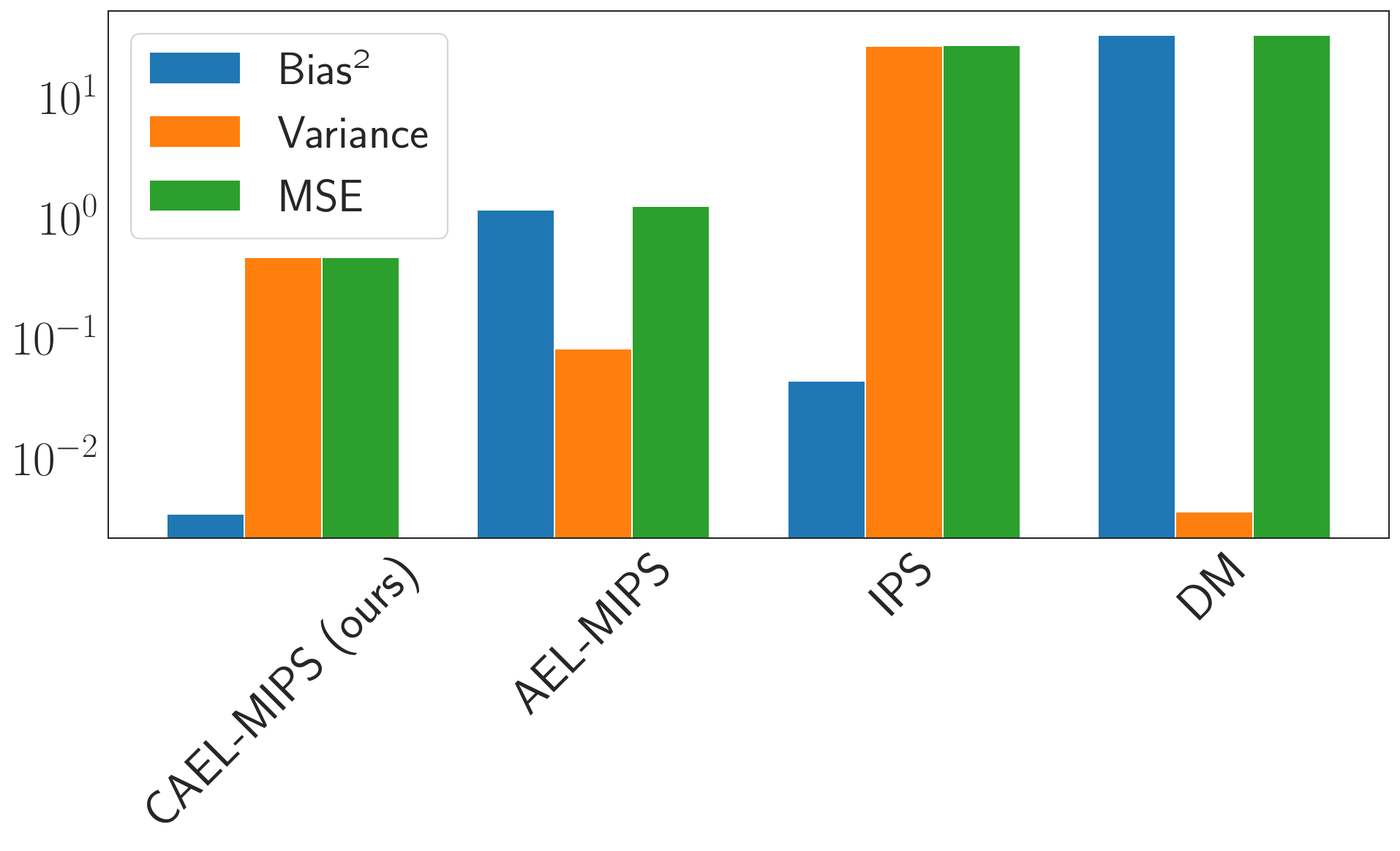}
    \caption{Bias, variance, and MSE comparison between \method, AEL-MIPS, IPS, and DM (AEL-MIPS refers to the method of \citet{cief2024learning}). The vertical axis is plotted on the log scale. The values are estimated by using 1000 samples and averaging over 30 independent runs.}
    \label{fig:bias-var-mse}
    \Description[]{Bias, variance, and MSE of different estimators for the synthetic dataset.}
\end{figure}

For experiments, \method is tested on both a synthetic dataset and a real-world dataset. We prepared the synthetic dataset to test OPE methods on various settings and to empirically validate the theoretical findings in \cref{sec:method}. We also test \method on a real-world bandit dataset \citep{saito2022off} to demonstrate the applicability of \method in practical scenarios and to show that \method outperforms baselines in most experimental runs. We use $\alpha = 10$ and $\beta = 0.1$ for \method in all the experiments. The baselines used in our experiments include AEL-MIPS \citep{cief2024learning}, IPS, and DM. AEL-MIPS is included as a baseline, as it learns embeddings for MIPS, similar to \method, but only learns embeddings for actions based on the reward prediction loss (\cref{eq:emp-r-loss}). We also compare our proposed model with DM, as both \method and AEL-MIPS utilize the reward prediction loss for learning embeddings. For DM, we estimate the policy value using the approximated expected reward learned through the reward prediction loss: 
\begin{align}\label{eq:dm}
    \hat v_{\text{DM}}(\pi) = \frac1n \sum_{i=1}^n \sum_{a \in \cA} \pi(a|x_i) \hat q(x_i,a) \,.
\end{align}
where $\hat q$ is the learned reward model. Lastly, we compare with IPS to see if MIPS methods with embedding learning reduce the MSE of the OPE estimation.

\subsection{Synthetic Data}\label{sec:synthetic-exp}


We conduct a synthetic data experiment to assess the capacity of \method on reducing the MSE of MIPS under diverse experimental conditions. The subsequent description outlines the sampling distributions used for rewards, actions, and contexts in the synthetic data, along with a description of the target policy used in the experiment.



\paragraph{\textbf{Rewards}}
The reward distribution for the synthetic data is  $P_R=\cN(q(x,a), 1)$, where $q(x,a) = 10 e^{-(x_1 - a_1)^2}$, and $x_1$ and $a_1$ are the first components of the context and action vectors respectively. Since AEL-MIPS employs a linear reward model for embedding learning, we construct the reward function $q(x,a)$ to be a nonlinear function of context–action pairs to examine whether \method, which does not utilize a linear reward function, can achieve lower MSE than AEL-MIPS.

\paragraph{\textbf{Actions}}
The action space consists of $k=500$ actions. Each action $a$ has a vector representation where its first component is $a/k$, and the remaining $k-1$ components are sampled uniformly at random from $[0, 1]$. The actions in the dataset were sampled from a uniform random behavior policy unless specifed otherwise. 

\paragraph{\textbf{Contexts}}
The contexts in the synthetic data were sampled uniformly from the context space $\cX = [0, 1]^5$.

\paragraph{\textbf{Target Policy}}
The target policy $\pi$ is $\epsilon$-greedy with respect to $q(x,a)$ with $\epsilon = 0.2$.

Using the synthetic data, we first compare \method and baselines in terms of bias, variance, and MSE (\cref{fig:bias-var-mse}). \method achieves the lowest MSE among the OPE estimators. \method has the bias of OPE estimation significantly reduced while inducing a small increase in the variance compared to AEL-MIPS. This is due to the additional MSE-minimizing loss terms in the learning objective of \method in addition to the reward prediction loss used in AEL-MIPS. The use of a nonlinear reward model further contributes to the bias reduction, as AEL-MIPS assumes a linear reward model for embedding learning. Lastly, the context-action dependent embeddings of \method also contribute to a reduction in bias compared to AEL-MIPS. Both \method and AEL-MIPS show significantly reduced MSE compared to IPS due to substantially reduced variance through learned embeddings. DM, which uses a reward function learned with the reward prediction loss for OPE estimation, exhibits the highest bias and the lowest variance among OPE methods. The bias arises from the reward model overfitting to uniformly sampled, uninformative action components, while the low variance results from the deterministic mapping between predicted rewards and context–action pairs.

\begin{figure*}[t]
    \centering
   \begin{minipage}{0.8\textwidth}
       \centering
       \includegraphics[width=0.8\linewidth]{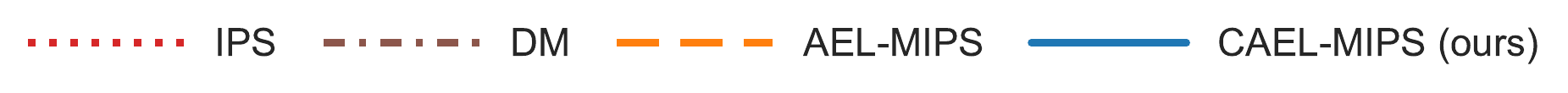}
   \end{minipage}

    \begin{subfigure}{0.3\textwidth}
        \centering
        \includegraphics[width=\linewidth]{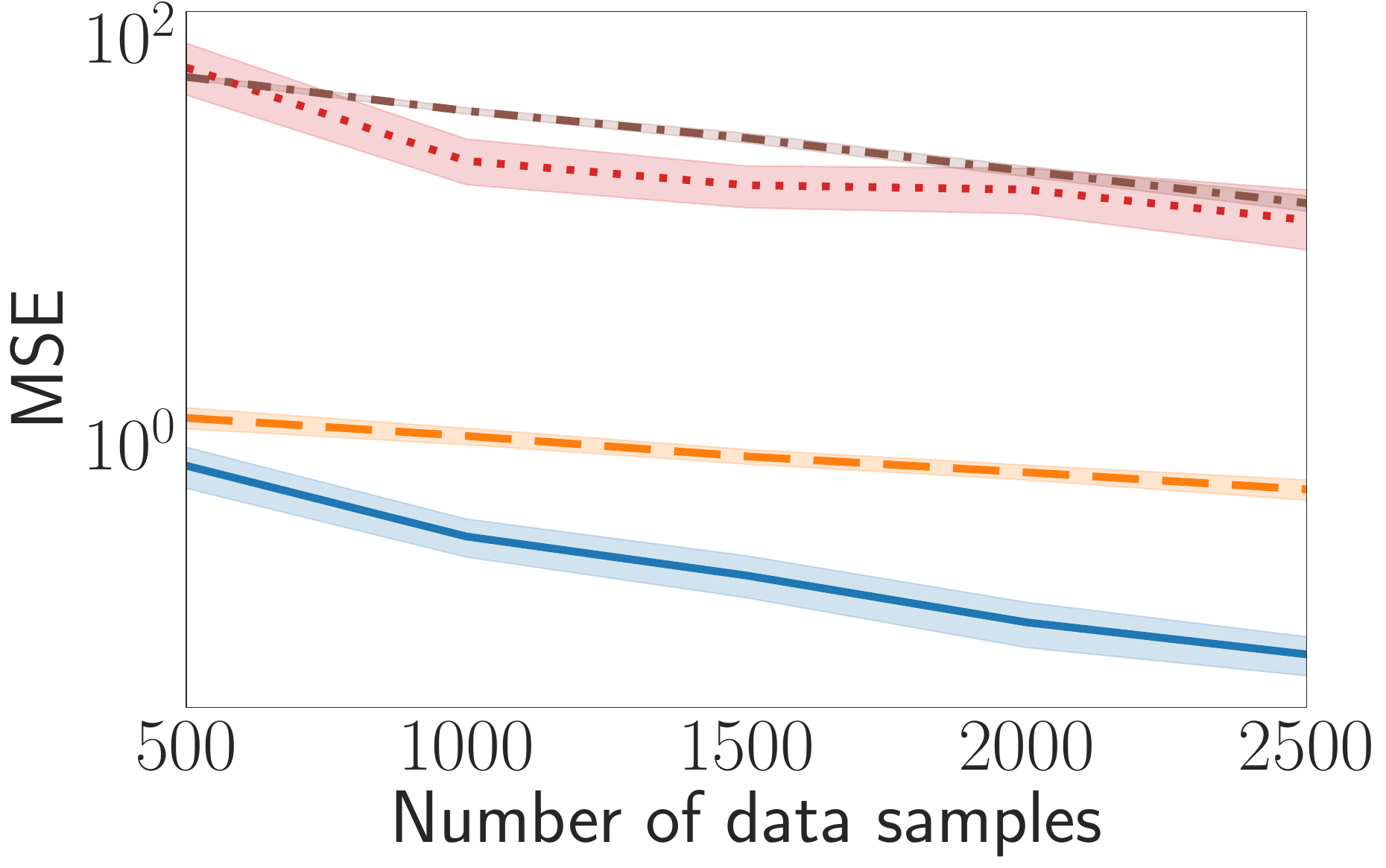}
        \caption{}
        \label{fig:n_val}
    \end{subfigure}\hfill
    \begin{subfigure}{0.3\textwidth}
        \centering
        \includegraphics[width=\linewidth]{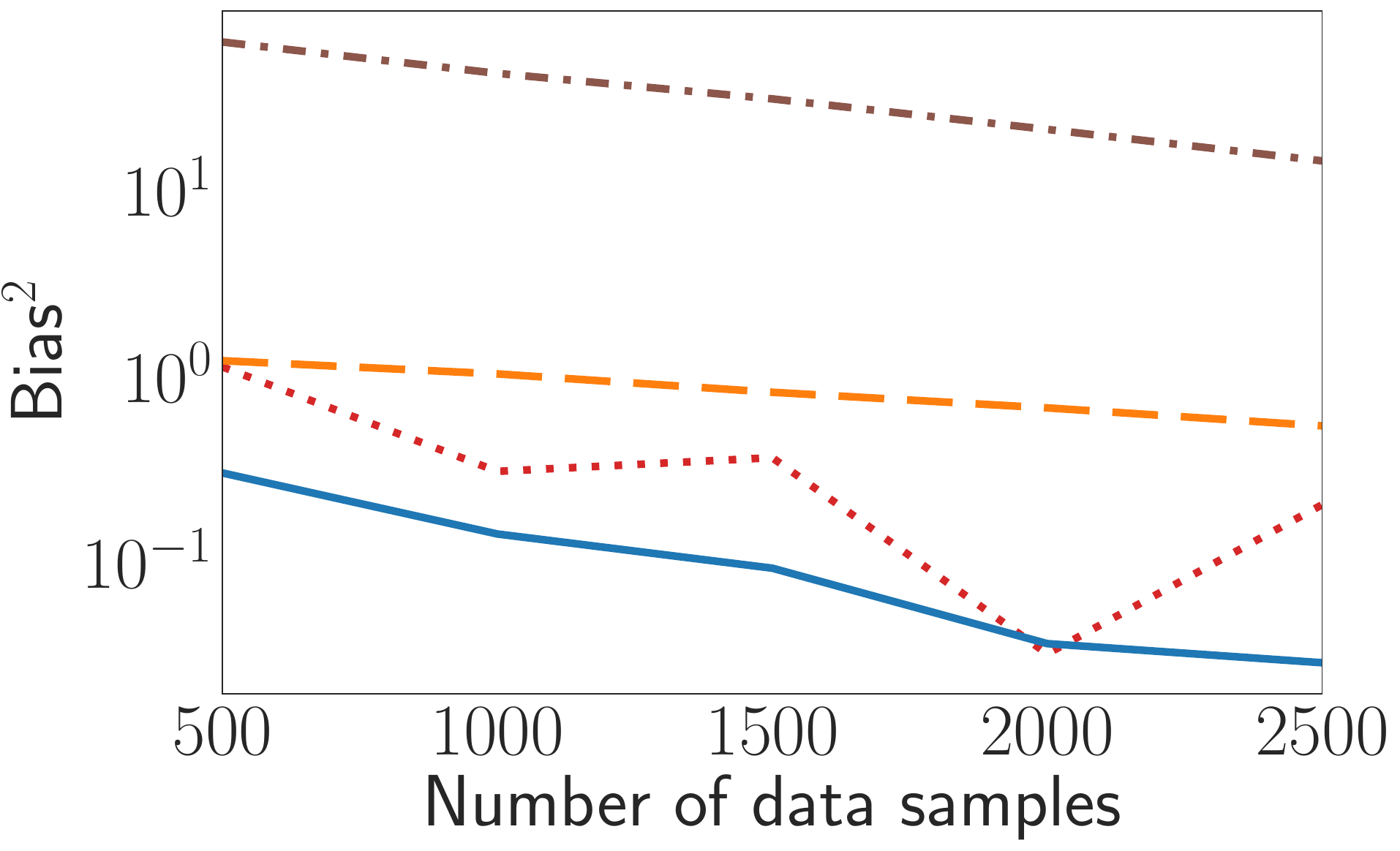}
        \caption{}
        \label{fig:n_val_bias}
    \end{subfigure}\hfill
    \begin{subfigure}{0.3\textwidth}
        \centering
        \includegraphics[width=\linewidth]{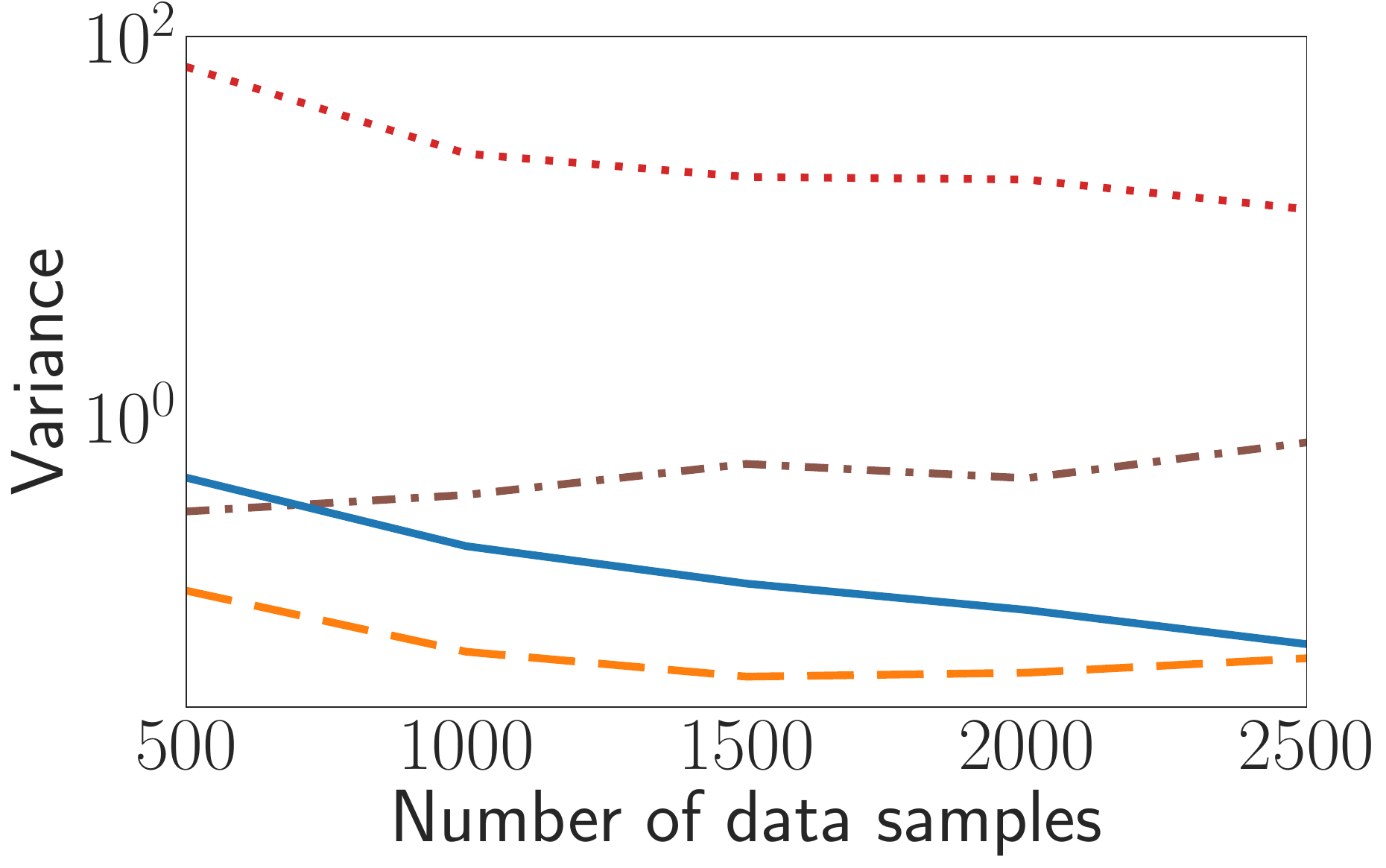}
        \caption{}
        \label{fig:n_val_var}
    \end{subfigure}
   
    \begin{subfigure}{0.3\textwidth}
        \centering
        \includegraphics[width=\linewidth]{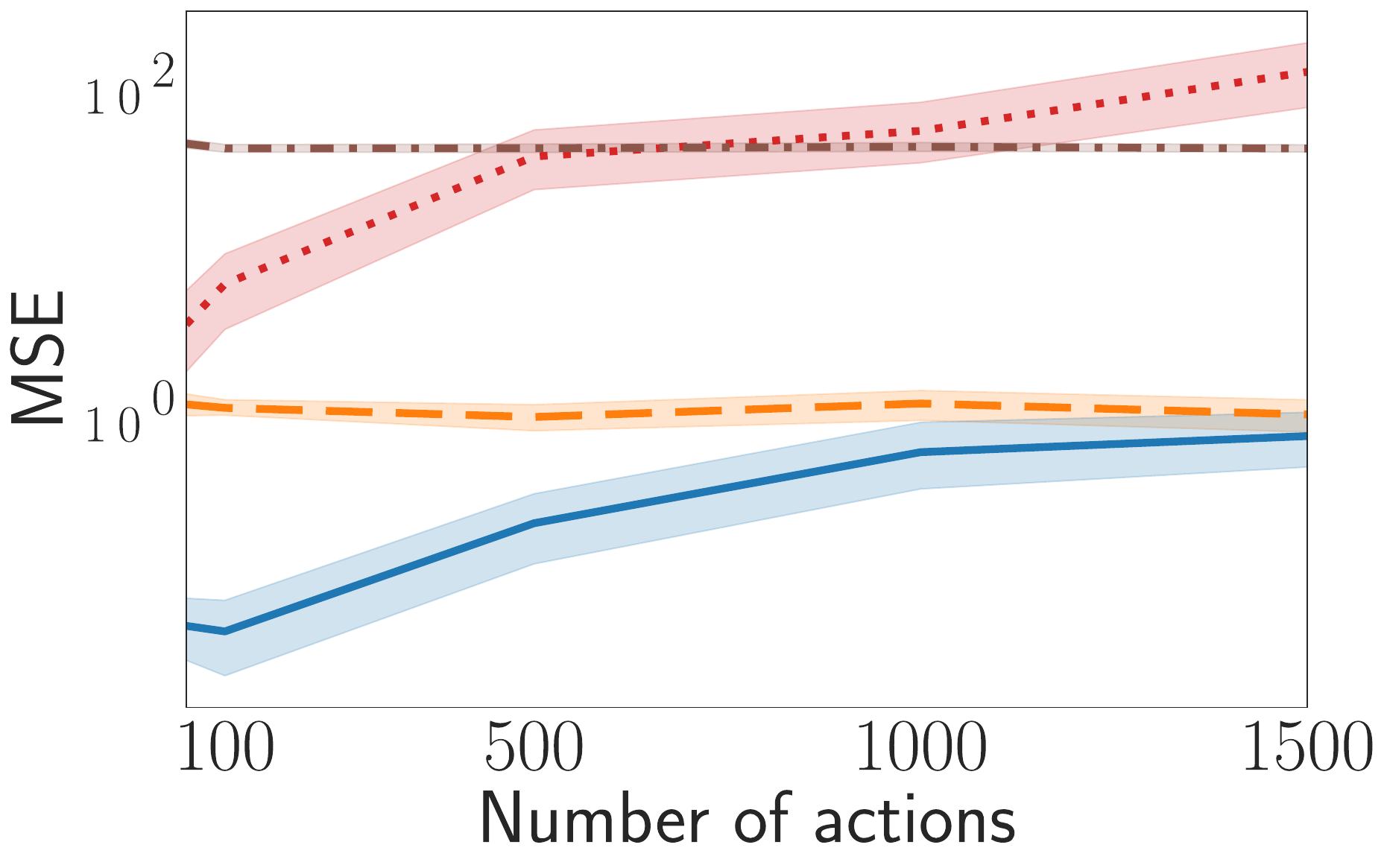}
        \caption{}
        \label{fig:n_act}
    \end{subfigure}\hfill
    \begin{subfigure}{0.3\textwidth}
        \centering
        \includegraphics[width=\linewidth]{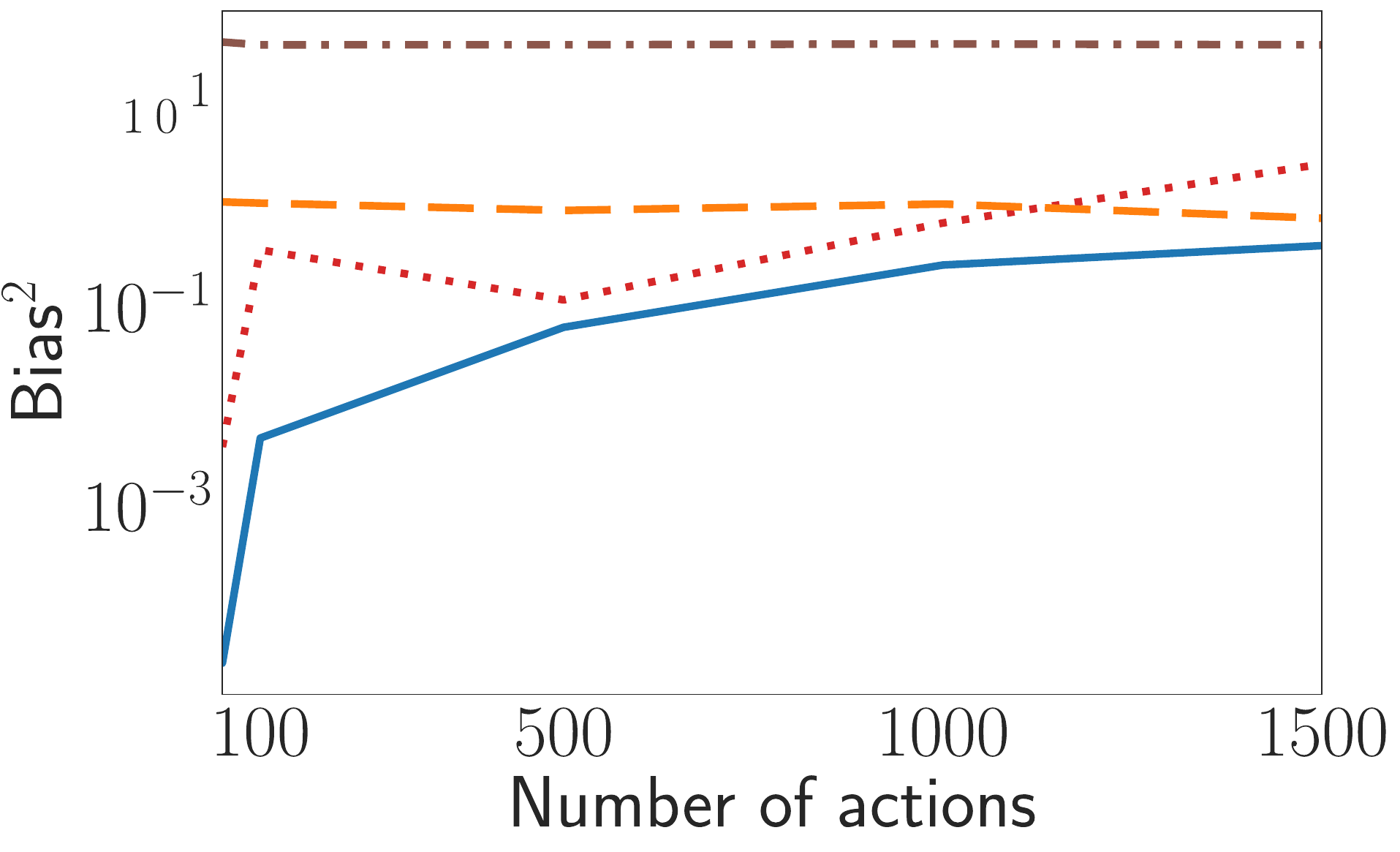}
        \caption{}
        \label{fig:n_act_bias}
    \end{subfigure}\hfill
    \begin{subfigure}{0.3\textwidth}
        \centering
        \includegraphics[width=\linewidth]{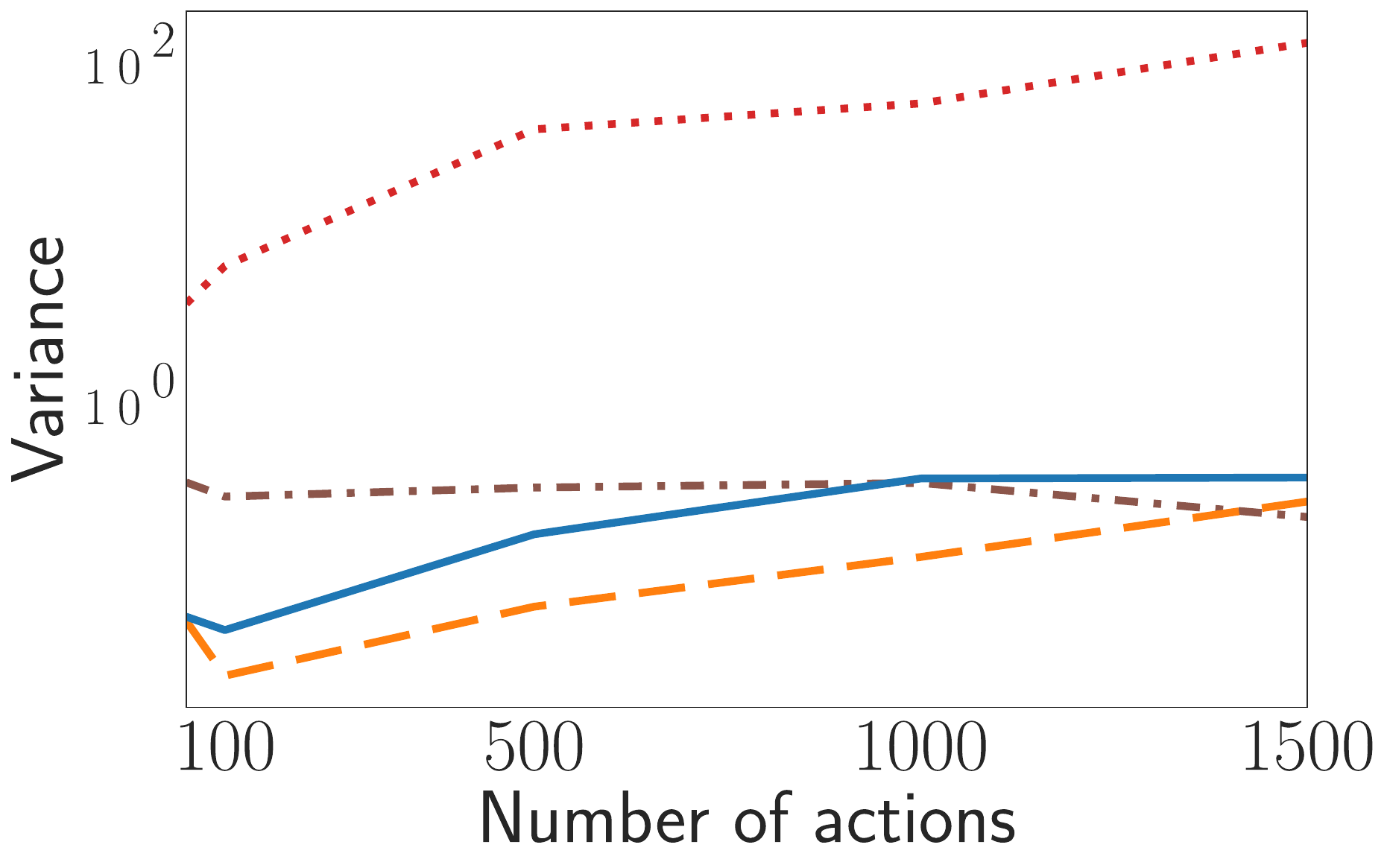}
        \caption{}
        \label{fig:n_act_var}
    \end{subfigure}

    \begin{subfigure}{0.3\textwidth}
        \centering
        \includegraphics[width=\linewidth]{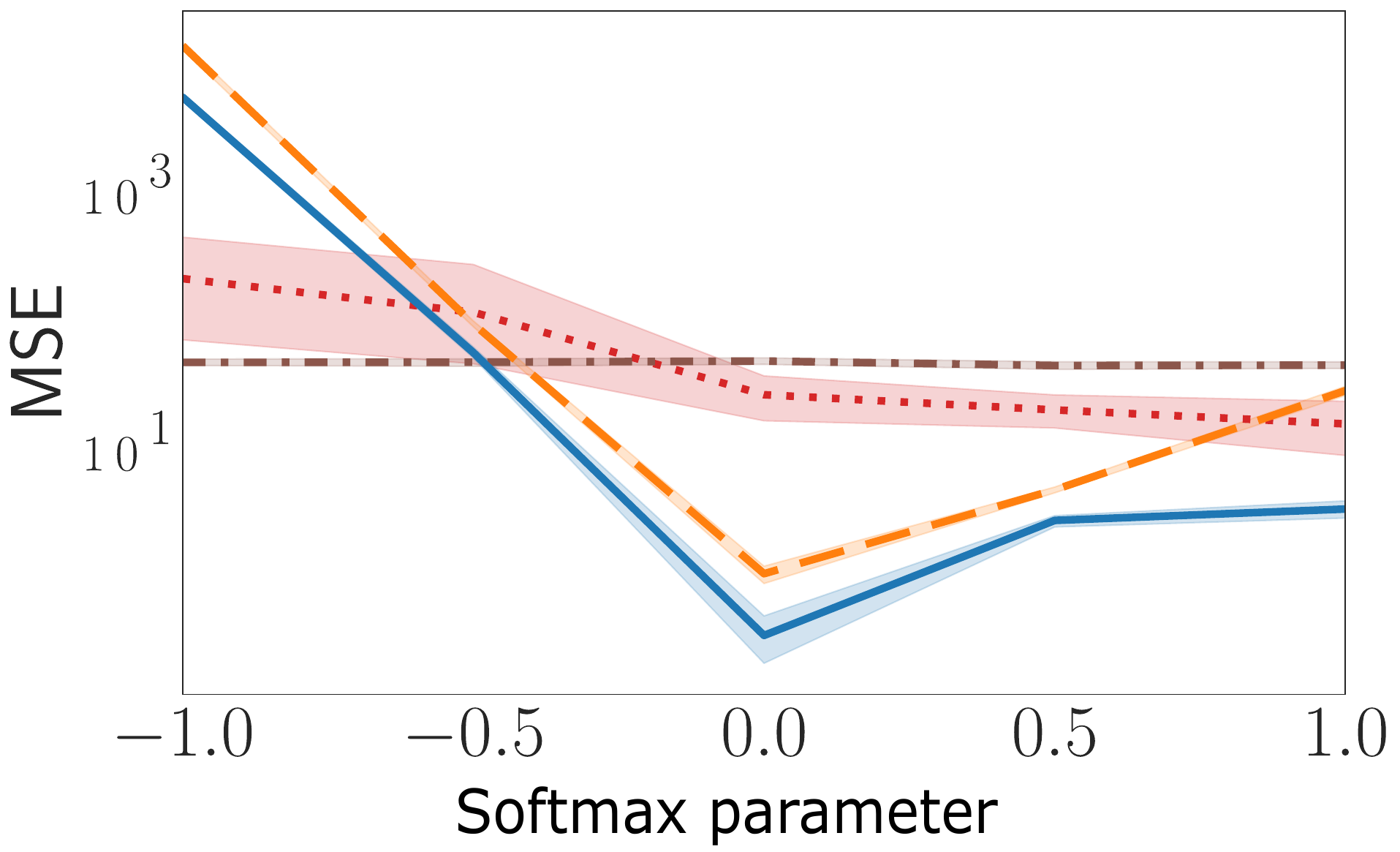}
        \caption{}
        \label{fig:beta}
    \end{subfigure}\hfill
    \begin{subfigure}{0.3\textwidth}
        \centering
        \includegraphics[width=\linewidth]{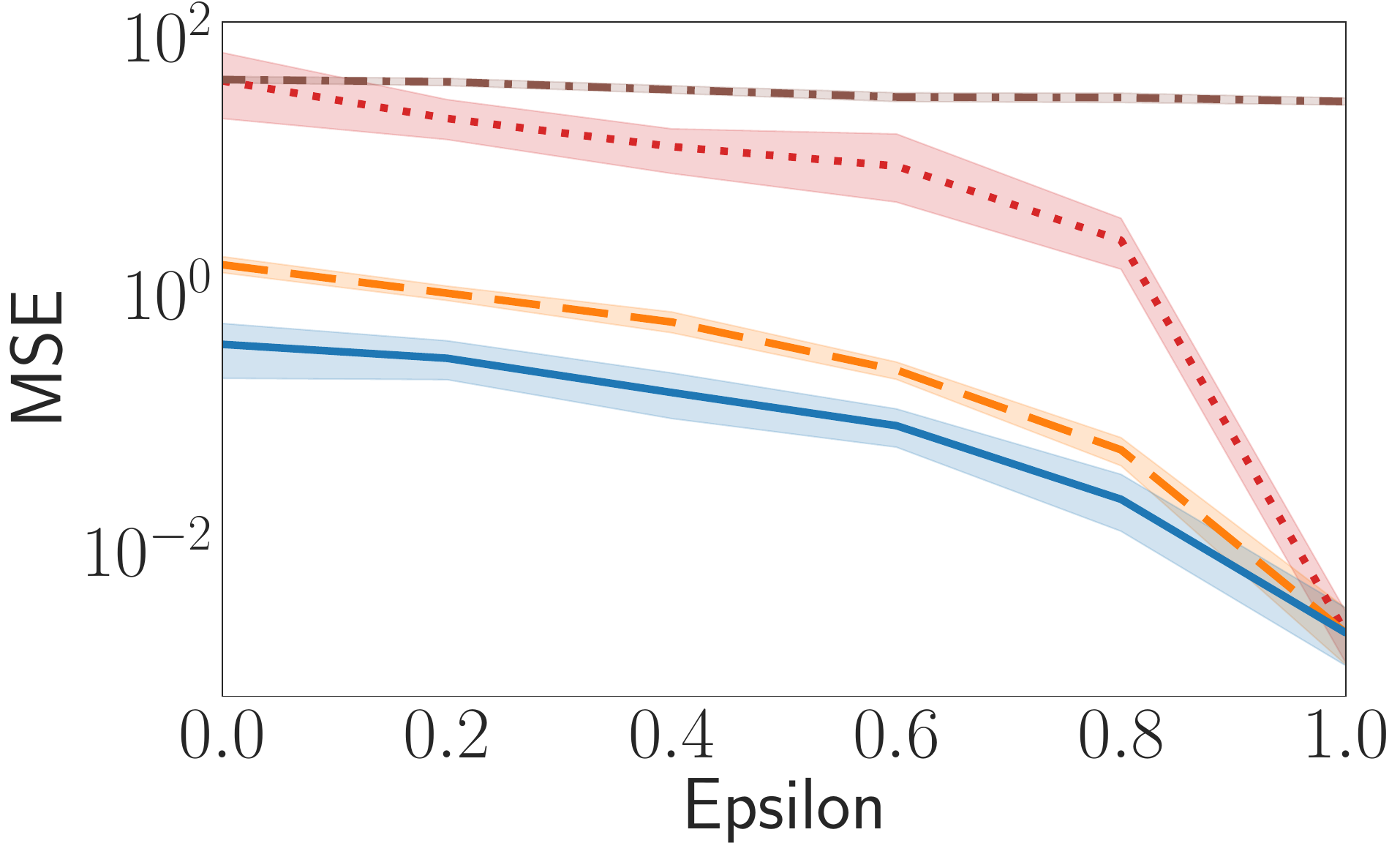}
        \caption{}
        \label{fig:eps}
    \end{subfigure}\hfill
    \begin{subfigure}{0.3\textwidth}
        \centering
        \includegraphics[width=\linewidth]{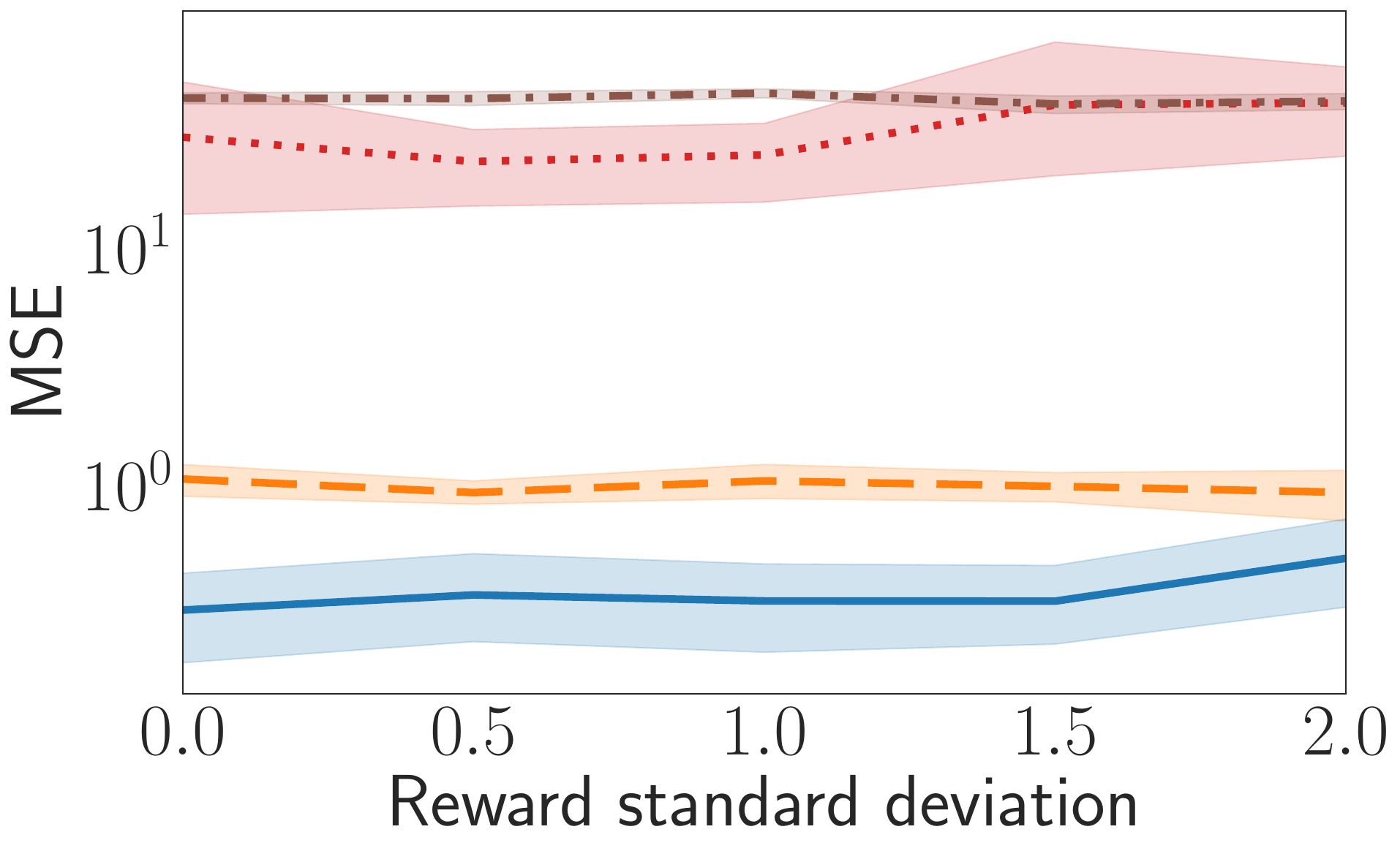}
        \caption{}
        \label{fig:r_std}
    \end{subfigure}
    
    \caption{MSE, bias, and variance of IPS, DM, AEL-MIPS, and \method as a function of various problem parameters for the synthetic dataset. For all the plots, the vertical axis is plotted on the log scale and the shaded regions represent 95\% confidence interval. \cref{fig:n_val,fig:n_val_bias,fig:n_val_var} show the MSE, bias and variance as the number of samples in the dataset increases. \cref{fig:n_act,fig:n_act_bias,fig:n_act_var} show the MSE, bias, and variance as the number of actions increases. \cref{fig:beta,fig:eps,fig:r_std} shows the MSE of all the estimators as a function of various environment parameters, where softmax parameter is the inverse temperature parameter of the softmax behavior policy, and Epsilon is the suboptimality of the target policy. We observe that the trend for MIPS-based estimators is similar to IPS but with a larger reduction in MSE due to variance reduction. Furthermore, \method has the lowest MSE across different environment settings since it learns embeddings that balance bias-variance trade-off.}
    \label{fig:synthetic2}
    \Description[]{MSE as a function of different parameters.}
\end{figure*}

Next, we evaluate the performance of \method and baselines for various experimental conditions and see if the performance ordering observed in \cref{fig:bias-var-mse} persists (\cref{fig:synthetic2}). In \cref{fig:n_val,fig:n_val_bias,fig:n_val_var}, we show MSE of OPE estimators, as well as bias and variance that compose the MSE on a varying dataset size. The figures were drawn with 100 repeated trials for each dataset size. In \cref{fig:n_val}, we can observe that the performance ordering of OPE methods that was observed in \cref{fig:bias-var-mse} persists on varying dataset size. \method outperforms other OPE estimators in terms of MSE. \method shows significantly reduced bias while inducing a small variance compared to AEL-MIPS. However, as the dataset size becomes larger, \method reduces both bias and variance due to the MSE-minimizing loss terms in its embedding learning objective and the variance of \method becomes similar to the variance of AEL-MIPS. Both IPS and DM shows large MSE due to large variance and bias respectively as in \cref{fig:bias-var-mse}. 

\cref{fig:n_act,fig:n_act_bias,fig:n_act_var} show the MSE, bias, and variance of OPE estimators with respect to the number of actions. We test OPE estimators on increasing variance due to increaseing number of actions \citep{sachdeva2024off} and see if the performance ordering observed in \cref{fig:bias-var-mse} persists. The figures were drawn with 30 repeated trials for each number of actions (50, 100, 1000, 1500). In \cref{fig:n_act_var}, IPS, AEL-MIPS, and \method show increase in variance as the number of actions increases while DM is not significantly affected by the increase in the number of actions. For IPS that has variance as a dominant factor in its MSE, the increase in the variance is reflected heavily on its MSE. In \cref{fig:n_act_bias} we observe that \method has the lowest bias among all the estimators due to the MSE-minimization loss terms. The bias grows with the number of actions, and our empirical results match the rate given by \cref{prop:entropy-bias}. \cref{fig:n_act_var} shows how the variance increases as we increase the number of actions.

\cref{fig:beta} shows MSE of OPE esetimators on varying behavior policy $\mu(a|x)$. The figure was drawn with 30 repeated trials for each value of $\gamma \in \{-1.0, -0.5, 0.0, 0.5, 1.0 \}$ that controls the softmax behavior policy $\mu(a|x)$:
\begin{align*}
    \mu(a|x) = \frac{e^{\gamma q(x,a)}}{\sum_{b \in \cA} e^{\gamma q(x,b)}} \,.
\end{align*}
When $\gamma = 0$, the policy is a uniform random distribution, while large values of $\gamma$ makes the policy tend towards greedy, and negative $\gamma$ values makes the policy suboptimal. We observe that for negative $\gamma$ values, the MSE is large as the mismatch between the target policy and the behavior policy is large. We also observe that the MSE of \method is always smaller than that of AEL-MIPS and the performance ordering observed in \cref{fig:bias-var-mse} generally persists in this setting. 

\cref{fig:eps} shows the MSE as a function of $\epsilon$ which controls the randomness of the $\epsilon$-greedy target policy with uniform random behavior policy. The figure was drawn with 30 repeated trials for each value of $\epsilon$. When $\epsilon=0$, the target policy is greedy, and is the optimal policy. Since the behavior policy is uniform random, as the target policy changes from greedy to uniform random by increasing $\epsilon$ from zero to one, target policy approaches uniform random behavior policy and MSE approaches zero for all OPE estimators. However, the performance ordering of the OPE estimators observed in \cref{fig:bias-var-mse} is still observed in this experiment setting and \method outperforms the baseline OPE estimators across varying $\epsilon$.  

\cref{fig:r_std} shows the MSE on varying reward distribution. The figure was drawn with 30 repeated trials for each value of standard deviation that controls the reward distribution. Similar to the other empirical study made on varying experiment environment parameters, we can see that the performance ordering observed in \cref{fig:bias-var-mse} is still observed across varying standard deviation of the reward distribution.

\begin{figure}[t]
    \centering
    \includegraphics[width=\linewidth]{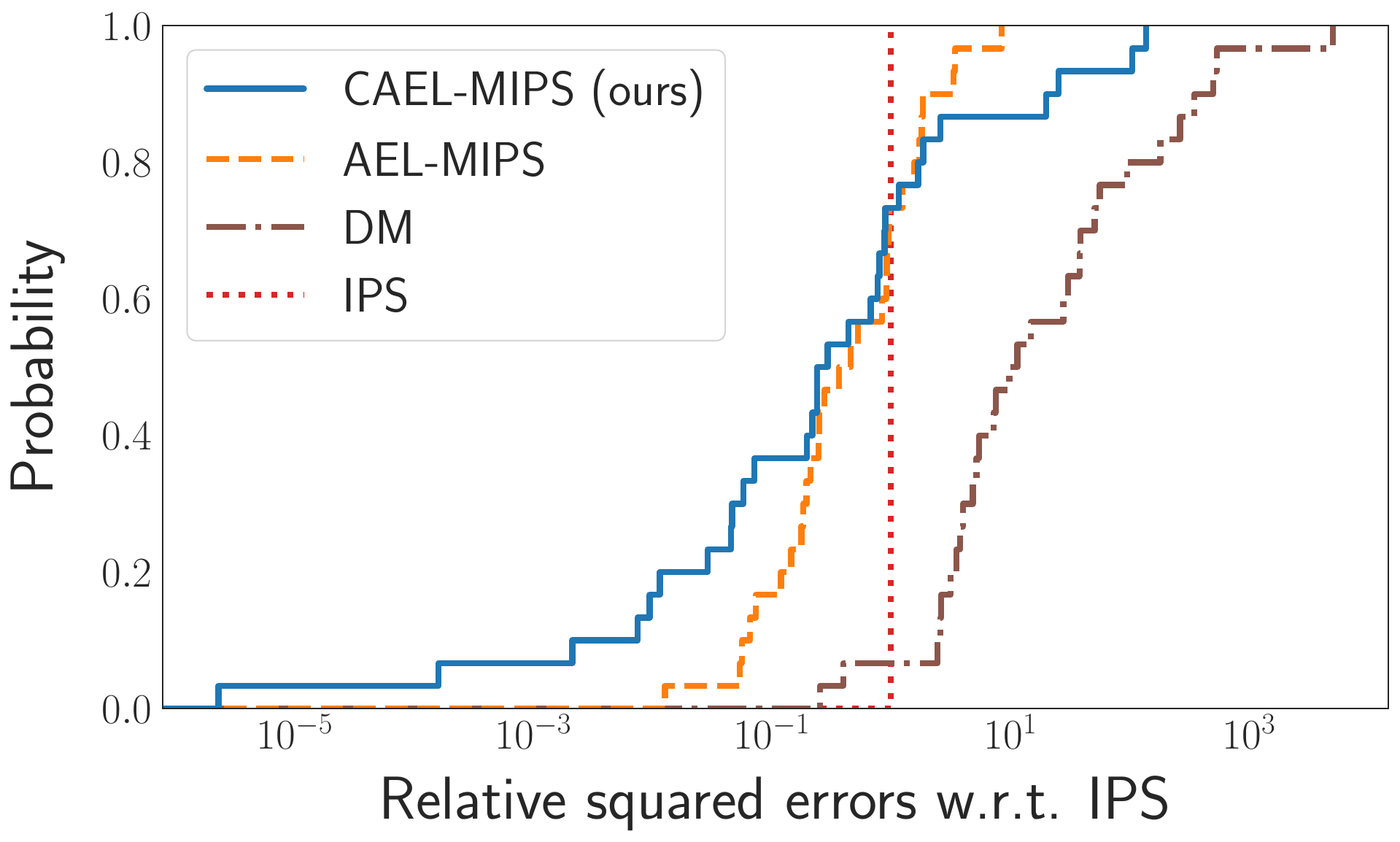}
    \caption{Empirical cumulative distribution function (CDF) of relative errors of different estimators for the real-world bandit dataset. The relative errors are computed by dividing the squared errors of all the estimators by the squared error of IPS.}
    \label{fig:obp}
    \Description[]{Empirical cumulative distribution function of relative errors of different estimators for the real-world dataset.}
\end{figure}

\subsection{Real-World Data}\label{sec:real-world-exp}
In this section, we test \method and baselines on a complex real-world dataset to assess the scalability of our method in practical use cases. We test the OPE estimators on \emph{Open Bandit Dataset}, a real-world bandit dataset collected on a fashion e-commerce platform ZOZOTOWN \citep{saito2020large}. The dataset was collected from the A/B test of two policies: a uniformly random behavior policy and a Thompson sampling target policy. We use $n=10000$ samples. The context dimension is $d=20$, the total number of actions is $|\cA| = 240$, and the reward is binary, which indicates whether the user clicked the recommended item or not. To learn the embeddings, we use the same three-layer feedforward neural network as discussed in \cref{sec:alg}. We use one-hot action representations as input to the neural network. For DM, we use the same network as \method, and use it as a reward model to estimate $v(\pi)$ using \cref{eq:dm}. We repeat the experiments 30 times to estimate the squared errors of all the estimators.

We plot the empirical cumulative distribution function (CDF) of the squared errors relative to IPS in \cref{fig:obp} following the work of \citet{saito2022off} and \citet{cief2024learning}. To compute the squared errors relative to IPS, the squared errors of all the estimators are normalized by dividing them by the squared error of IPS. Then empirical CDF of the relative squared errors is computed. From \cref{fig:obp}, we observe that \method outperforms IPS in more than 75\% of the runs. Furthermore, while the relative errors of AEL-MIPS and DM never reach below 0.1 and 0.01, respectively, \method's relative error reaches below 0.1 for about 40\% of the runs and below 0.01 for about 20\% of the runs.


\section{Discussion}\label{sec:discussion}
In this work, we proposed \method, a MIPS-based estimator that learns context-action embeddings to perform OPE in contextual bandits. We derived upper bounds for bias and variance of MIPS, and proposed an embedding learning objective function using those bounds and the reward prediction loss. By minimizing the MSE upperbound in the embedding learning objective, the learned embeddings directly reduce the MSE of MIPS estimator. In the theoretical analysis, we showed that our embedding learning objective function is related to the entropy of the posterior distribution $\mu(a|x,e)$. In our experiments, we empirically validated that \method outperforms the baselines on various experiment settings on synthetic dataset due to its context-action dependent embeddings learned through the learning objective that directly minimizes MSE. We further demonstrated the scalability of \method to practical use cases by testing it on a real-world bandit dataset, where it consistently outperformed the other baselines.

Our work opens several directions for future research. Due to the unknown degrees of tightness of the upper bounds on bias and variance of MIPS, our method introduced constant weights for bias and variance upperbounds used in the embedding learning objective, which may be difficult to tune. A potential future work is to choose these hyperparameters using methods such as SLOPE \citep{su2020doubly,tucker2021improved}. Furthermore, computing the bias upper bound in \cref{alg:main-algo} requires $O(|\cA|^2)$ operations. Reducing this computation would make our method more scalable to large action spaces \citep{lopez2021learning}. In addition, \cref{alg:main-algo} relies on multiple calls to a logistic regression subroutine to estimate the posterior $\hat \mu(\cdot | x, e)$. A potential future work would be to use a generative network to model the posterior. Finally, extending our method to policy optimization and the full reinforcement learning setting are also promising directions.





\bibliography{arlet_ope}

\appendix

\section{Proofs}
\label{appendix:proofs}

\begin{proof}[Proof of \cref{prop:entropy-bias}]
    We first show the upper bound on $\cL_\bias(\theta)$, that is, $\cL_\bias(\theta) \le (C^2/4)(1-\EE[e^{H_2(X,A;\theta)}])^2$. First recall from \cref{eq:bias-loss} that
    \begin{align*}
        &\cL_\bias(\theta) \\
        & = \EE\left[ \sum_{a<b} \mu(a|X,f_\theta(X,A)) \mu(b|X,f_\theta(X,A)) |w(X,b) - w(X,a)| \right]^2 \,.
    \end{align*}
    Since $|w(x,a) - w(x,b)| < C$ for all $(x,a,b) \in \cX \times \cA^2$, we have 
    \begin{align*}
        \cL_\bias(\theta) \le C^2 \EE\left[ \sum_{a<b} \mu(a|X,f_\theta(X,A)) \mu(b|X,f_\theta(X,A)) \right]^2 \,.
    \end{align*}
    It remains to bound $\sum_{a<b} \mu(i|X,f_\theta(X,A)) \mu(j|X,f_\theta(X,A))$. Recall the identity $2\sum_{a<b} a_i a_j = (\sum_i a_i)^2 - \sum_i a_i^2$. Applying it to the posterior $\mu(\cdot|x,f_\theta(x,a))$ for all $(x,a) \in \cX \times \cA$, we have
    \begin{align*}
        & \sum_{i<j} \mu(i|x,f_\theta(x,a)) \mu(j|x, f_\theta(x,a)) \\
        &= \frac12 \left( \left( \sum_{i \in \cA} \mu(i|x,f_\theta(x,a)) \right)^2 - \sum_{i \in \cA} \mu(i|x,f_\theta(x,a))^2 \right) \\
        &= \frac12 \left( 1 - \sum_{i \in \cA} \mu(i|x,f_\theta(x,a))^2 \right) \\
        &= \frac12 \left( 1 - e^{-H_2(x, a;\theta)} \right) \,,
    \end{align*}
    which proves the result. The second inequality follows from the fact that the entropy is maximized for the uniform distribution of the actions. Now, from Taylor series expansion, we have $e^{-x} \ge 1-x$ for all $x \in \R$. Applying it to $e^{-H_2(X,A;\theta)}$, we obtain 
    \[
        \cL_\bias(\theta) \le (C^2/4)\EE[H_2(X,A;\theta)] \,,
    \]
    proving the last inequality.
\end{proof}

\begin{proof}[Proof of \cref{prop:entropy-var}]
    Recall that 
    \begin{align*}
            \cL_{\text{var}}(\theta) = & \frac{1}{n} \EE\Bigg[ \EE[R\,|\,X,E]^2 \sum_{a \in \cA}  \mu (a|X, f_\theta(X, A))^2 \nonumber \\
    & \qquad \qquad\sum_{a \in \cA} w(a, X)^2 \Bigg]
    \end{align*}
    The proposition follows by noting that from the definition of $H_2$, we have $\sum_a \mu(a|X,f_\theta(X,A)) = e^{-H_2(X,A;\theta)}$. The upper bound can be obtained by noting that $H_2(X,A;\theta) \ge 0$, implying that $e^{-H_2(X,A;\theta)} \le 1$, the boundedness of IPS weights, and the fact that $\EE[R|X,E]^2$ does not depend on $a$.
\end{proof}

\begin{proof}[Proof of \cref{prop:bias-bound}]
    Our goal is to bound $|\Bias[\vmips(\pi)]$. From \cref{eq:mips-bias}, recall that the bias of MIPS is given by \citep{saito2022off}: 
    \begin{align*}
        & \Bias[\vmips(\pi)]  \\
        & = \EE\Biggl[ \sum_{a < b} \mu(a|X,E)\mu(b|X,E)  \\
        &  \qquad\qquad \cdot (q(X,a,E) - q(X,b,E)) \cdot (w(X,b) - w(X,a)) \Biggr]  \,.
    \end{align*}
    Now, we have that
    \begin{align*}
    & |\Bias[\vmips(\pi)]| \le \EE\Biggl[ \Biggl\lvert \sum_{a < b}\mu(a|X,E)\mu(b|X,E) \\
    & \;\;\;\;\;\;\;\;\;\;\cdot (q(X,a,E) - q(X,b,E)) \cdot  (w(X,b) - w(X,a)) \Biggr\rvert \Biggr] \tag{$\clubsuit$} \\
    &\le \EE\Biggl[ \sum_{a<b} \abs{ \mu(a|X,E)\mu(b|X,E) } \\
    & \;\;\;\;\;\;\;\;\;\;\cdot \abs{ (q(X,a,E) - q(X,b,E)) }  \cdot \abs{ (w(X,b) - w(X,a)) } \Biggr] \tag{$\spadesuit$} \\
    &\le (q_{\max}-q_{\min})\\
     &\;\;\;\;\;\;\;\;\;\;\cdot \EE\Biggl[ \sum_{a<b} \mu(a|X,E)\mu(b|X,E) \abs{w(X,b) - w(X,a)} \Biggr] \,, \tag{$\diamondsuit$} 
\end{align*}
where we used Jensen's inequality in $(\clubsuit)$, triangle inequality in $(\spadesuit)$, and \cref{as:bounded-mean-var} in $(\diamondsuit)$.
\end{proof}

\begin{proof}[Proof of \cref{prop:var-bound}]
Here, we upper bound the variance of MIPS. Recall from \cref{eq:mips-var} that
\begin{align}\label{eq:mips-var}
    \Var[\vmips(\pi)] &=  \Var[\vips(\pi)]  - \frac1n\EE[\EE_{P_{R|X,E}}[R^2]\Var_{\mu(\cdot|X,E)}[w(X,A)]] \nonumber\\
    &= \Var[\vips(\pi)] - \frac1n\EE\Bigg[ \left(\Var[R|X,E] + \EE[R|X,E]^2 \right) \nonumber \\
    & \qquad \qquad \qquad \qquad \Var_{\mu(\cdot|X,E)}[w(X,A)] \Bigg] \nonumber\\
    &\le \Var[\vips(\pi)] - \frac1n\EE\Bigg[ \left(\sigma_{\min}^2 + \EE[R|X,E]^2 \right) \nonumber \\
    & \qquad \qquad \qquad \qquad \Var_{\mu(\cdot|X,E)}[w(X,A)] \Bigg] \,,
\end{align}
where the last inequality follows from \cref{as:bounded-mean-var}. We now upper bound the RHS of \cref{eq:mips-var} as follows.
\begin{align}\label[ineq]{ineq:mips-var-ub}
    & \Var[\vips(\pi)] - \frac1n\EE\Bigg[ \left(\sigma_{\min}^2 + \EE[R|X,E]^2 \right) \Var_{\mu(\cdot|X,E)}[w(X,A)] \Bigg] \nonumber \\
    &= \Var[\vips(\pi)] - \frac1n\EE\Bigg[ \left(\sigma_{\min}^2 + \EE[R|X,E]^2 \right) \nonumber \\
    & \qquad \qquad \qquad \left( \EE_{\mu(\cdot|X,E)}[w(X,A)^2] - \EE_{\mu(\cdot|X,E)}[w(X,A)]^2 \right) \Bigg] \nonumber \\
    &= \Var[\vips(\pi)] + \frac{1}{n} \EE\left[\left( \sigma_{\min}^2 + \EE[R|X,E]^2 \right) \EE_{\mu(\cdot|X,E)}\left[ w(X,A) \right]^2 \right] \nonumber\\ 
    & \qquad \qquad - \frac{1}{n} \EE\left[ \left( \sigma_{\min}^2 + \EE[R|X,E]^2 \right) \EE_{\mu(\cdot|X,E)}\left[w(X,A)^2 \right] \right] \nonumber \\
    & \le \Var[\vips(\pi)] + \frac{1}{n} \EE\left[\left( \sigma_{\min}^2 + \EE[R|X,E]^2 \right) \EE_{\mu(\cdot|X,E)}\left[ w(X,A) \right]^2 \right] \nonumber \\
    &= \Var[\vips(\pi)] + \frac{1}{n} \EE\Bigg[ \left( \sigma_{\min}^2 + \EE[R|X,E]^2 \right) \nonumber \\
    & \qquad \qquad \qquad \qquad \qquad  \left( \sum_a \mu(a|X,E) w(X,a) \right)^2 \Bigg] \nonumber \\
    & \le \Var[\vips(\pi)] + \frac{1}{n} \EE\Bigg[\left( \sigma_{\min}^2 + \EE[R|X,E]^2 \right) \nonumber \\
    & \qquad \qquad \qquad \qquad \qquad \sum_a \mu(a|X,E)^2 \sum_a w(X,a)^2 \Bigg]
\end{align}
where we use Cauchy-Schwarz in the final step. Combining \cref{eq:mips-var} and \cref{ineq:mips-var-ub}, we get
\begin{align*}
    & \Var[\vmips(\pi)] \le \Var[\vips(\pi)] \\
    & + \frac{1}{n} \EE\Bigg[\left( \sigma_{\min}^2 + \EE[R|X,E]^2 \right) \sum_a \mu(a|X,E)^2 \sum_a w(X,a)^2 \Bigg]
\end{align*}
\end{proof}

\bibliographystyle{ACM-Reference-Format}
\end{document}